# Learning Bayesian networks from demographic and health survey data

Neville Kenneth Kitson[1, 2] and Anthony C. Constantinou[1, 3]

1. Bayesian Artificial Intelligence research lab, Risk and Information Management (RIM) research group, School of Electronic Engineering and Computer Science, Queen Mary University of London (QMUL), London, UK, E1 4NS.
2. OneWorld UK, CAN Mezzanine, London, UK, SE1 4YR.
3. The Alan Turing Institute, British Library, 96 Euston Road, London, UK, NW1 2DB.

E-mail addresses: ken.kitson@oneworld.org (N. Kitson), a.constantinou@qmul.ac.uk (A. Constantinou)

**ABSTRACT:** Child mortality from preventable diseases such as pneumonia and diarrhoea in low and middle-income countries remains a serious global challenge. We combine knowledge with available Demographic and Health Survey (DHS) data from India, to construct Causal Bayesian Networks (CBNs) and investigate the factors associated with childhood diarrhoea. We make use of freeware tools to learn the graphical structure of the DHS data with score-based, constraint-based, and hybrid structure learning algorithms. We investigate the effect of missing values, sample size, and knowledge-based constraints on each of the structure learning algorithms and assess their accuracy with multiple scoring functions. Weaknesses in the survey methodology and data available, as well as the variability in the CBNs generated by the different algorithms, mean that it is not possible to learn a definitive CBN from data. However, knowledge-based constraints are found to be useful in reducing the variation in the graphs produced by the different algorithms, and produce graphs which are more reflective of the likely influential relationships in the data. Furthermore, valuable insights are gained into the performance and characteristics of the structure learning algorithms. Two score-based algorithms in particular, TABU and FGES, demonstrate many desirable qualities; a) with sufficient data, they produce a graph which is similar to the reference graph, b) they are relatively insensitive to missing values, and c) behave well with knowledge-based constraints. The results provide a basis for further investigation of the DHS data and for a deeper understanding of the behaviour of the structure learning algorithms when applied to real-world settings.

## 1. Introduction

Preventing the deaths of children aged below 5 years in middle and low-income countries remains one of the world's major challenges. Despite recent progress, the United Nations International Children's Emergency Fund (UNICEF) estimate that 5.4 million children aged under 5 died in 2017 (UNICEF, 2018). The World Health Organisation (WHO) reports that the leading causes of death are from pre-term birth and intrapartum-related complications, respiratory infections, congenital abnormalities and diarrhoea (WHO, 2019). The majority of these deaths are preventable. Understanding the factors causing these deaths, and intervening to prevent them, is of great interest.

Over the past thirty years, the United States Agency for International Development (USAID) has "*pioneered the Demographic and Health Survey (DHS) Program ... to collect and share key information about people, their health, and their health systems*" (USAID, 2018). These surveys provide a dataset to which machine learning can be applied to explore the factors behind child mortality. The DHS Program has conducted over 400 surveys of demographic and health information across ninety low and middle-income countries, and the survey data is publicly available (ICF International, no date).

Causal Bayesian Networks (CBNs) model variables under causal assumptions and thus, can be used to model the impact of interventions available to decision makers. Correct identification of interventions is paramount in the area of child mortality as they have life-saving impacts, in addition to the need to identify cost-effective approaches to decision making. Moreover, CBNs are more readily interpretable than most other machine learning approaches. The visual graph produced shows the causal relationships between *all* the variables – for instance, which variables have a direct causal relationship on which others. This is particularly useful in the context of child mortality, since the relationships between contributing factors – such as wealth, education, breastfeeding, family planning practices - are themselves of interest. This ability to expose the relationships between all variables stands in contrast to other machine learning approaches which tend to be "blackbox" solutions where the focus is typically on maximising predictive accuracy of one target variable (e.g. child mortality), and causality is not explored.

While the relationships between variables encapsulated in CBNs can be specified by a human expert, this is often time-consuming and requires access to expertise which can also often be expensive. Alternatively, machine learning can be used to learn the graphical structure of the CBN from the observed data, potentially avoiding human bias. This paper explores this direction and assesses the usefulness of the DHS data as a basis for constructing a CBN model of the causes of childhood diarrhoea. The paper is structured as follows: Section 2 covers related work, Section 3 describes the pre-processing of available data to make it suitable for CBN structure learning, Section 4 presents the methodology, Section 5 discusses the results, and we provide our concluding remarks in Section 6.

## 2. Related work

### 2.1. Diarrhoea prevention and the factors which support or inhibit interventions

A joint report by UNICEF and WHO (2013) classifies interventions according to whether they protect against, prevent or treat diarrhoea as shown in Table 1. The table summarises the key interventions relating to diarrhoea, and an indication of the impact they have reducing diarrhoea prevalence or child mortality.

A range of demographic and socio-economic factors is identified in the literature which either support or inhibit these interventions. These indirect influences on the prevalence of diarrhoea include:

   i.  intervention cost and the economic resources of the family (Aunger et al., 2010; Dobe, Mandal and Jha, 2013),
   ii.  health awareness, media exposure, and education e.g. the effect of a promotional campaign on handwashing (Schmidt et al., 2009),
   iii.  cultural factors (Curtis, Danquah and Aunger, 2009),
   iv.  maternal education and household size (Dobe, Mandal and Jha, 2013),
   v.  geographic location (Fewtrell et al., 2005), and
   vi.  nutritional status of the child (Luby et al., 2009)

**Table 1.** Key interventions for prevention of diarrhoea - adapted from Table 1 in (UNICEF / WHO, 2013)

| Intervention type | Intervention | Effect |
|---|---|---|
| Protection | • Exclusive breastfeeding for 6 months | → Not breastfeeding in 0-5 months increased risk ratio for diarrhoea between 1.26 and 2.65 times |
| | • Continued breastfeeding from 6 - 23 months | → A risk ratio of 2.07 of diarrhoea incidence in infants from 6 - 23 months from not breastfeeding |
| | • Adequate complementary feeding from 6 - 23 months | → 6% reduction in all child deaths |
| | • Vitamin A supplementation | → 23% reduction in all-cause mortality |
| Prevention | • Vaccinations | → 74% reduction in very serious rotavirus diarrhoea infection |
| | • Handwashing with soap | → 31% and 48% diarrhoea risk reduction |
| | • Improved sanitation | → 36% diarrhoea risk reduction |
| | • Increased quantity of water | → 17% diarrhoea risk reduction with an advised 25 litres of water per day |
| | • Household water treatment and safe storage | → 31 - 52% diarrhoea risk reduction |
| Treatment | • Community-based care of diarrhoea | → Community care with Zinc and ORS reduced diarrhoeal deaths by 93% |
| | • Oral Rehydration Salts (ORS) | → Reduces diarrhoea mortality by 69% |
| | • Zinc | → Reduces diarrhoea mortality by 23% |

### 2.2. Past applications of Machine Learning to DHS data

The majority of papers using DHS data are based on mainstream statistical techniques such as multivariate logistical regression. The most relevant past studies we discovered include the examination of the association of feeding and hygiene with diarrhoea in Nepal by Acharya et al. (2018), and a study by Seid and Kelkay (2018) that looks at a broader range of factors related to diarrhoea in Nigeria. Likewise, Gebru et al. (2019) used multivariate logistic regression analysis on DHS data to look at the factors affecting stunting in Ethiopia.

Papers using machine learning to explore or model DHS data are much rarer. The DHS program's own article search page (ICF International, 2019) provided only three citations in response to a "machine learning" search term. In one of these, Khare et al. (2017) identified factors relevant to malnutrition using feature selection and then used decision trees to compare the relative importance of individual and community-level factors. Merzouki et al. (2019) used DHS data from 29 countries to relate socio-behavioural characteristics to HIV incidence. Principal Component Analysis was employed to identify key socio-behavioural characteristics related to HIV, and hierarchical clustering was then used to group countries using these characteristics. Xie et al. (2016) combined DHS data and satellite imagery to predict poverty across Africa based on night time light emission. We only found one relevant paper that investigates the application of CBNs with DHS data, by Nguefack-Tsague (2011) who examined a simple four-node manually constructed CBN model and showed that it compared favourably to linear regression models.

### 2.3. Learning Bayesian Networks

A Bayesian Network (BN) is a probabilistic graphical model proposed by Pearl (1982) and Kim and Pearl (1983). The graph of a BN is a Directed Acyclic Graph (DAG) and consists of nodes that represent variables and arcs between nodes that represent direct dependencies. The magnitude of the relationship between variables is specified in the so-called Conditional Probability Tables (CPTs).

The DAG represents the dependencies and conditional independencies present in the underlying probability distribution of the variables. In complex BNs with multiple connecting paths between nodes, one must consider conditional independence given one or more other nodes – known as the conditioning set. $D$-separation (Geiger et al., 1990) is used to determine whether two nodes are conditionally independent using information about the paths between them. This information includes whether the path includes nodes with two inbound arcs, such as $A \rightarrow B \leftarrow C$, known as v-structures that represent the unique causal class of conditional dependence, whereas A and C become dependent when conditioned on B; also known as the concept of *explaining away*.

While not all algorithms used in this study claim to discover causal structure, the following assumptions are made allowing us to interpret the BN as a causal BN – that is, the arcs show the direction of cause and effect:

- the causal Markov condition applies so that nodes are independent of their non-descendants conditional on their parents
- causal faithfulness – the independences and conditional independences entailed by the DAG using d-separation are the same as those present in the underlying probability distribution
- causal sufficiency – there are no unobserved variables that are common causes of two or more observed variables.

Conditional independencies in the data constrain the possible graphs and many algorithms use this information to learn the BN structure. These types of structure learning algorithms are called constraint-based and remove edges from the graph which are incompatible with the conditional independency relationships. They often use statistical tests to assess how likely the observed values are given the null hypothesis that the variables are indeed independent. If the $p$-value for the test is below a threshold (typically 0.01 or 0.05), the null hypothesis of independence is rejected and the variables are assumed to be dependent. Two independence tests are commonly used in structural learning algorithms include Pearson's Chi-Squared test (Pearson, 1900) and G-squared log-likelihood test (Woolf, 1957). Conditional independence is defined analogously, but with independence defined in the marginal contingency tables for each combination of conditioning values.

The conditional independencies in the data are not always sufficient to determine the directions of all the edges in the graph, which leads to a Partially Directed Acyclic Graph (PDAG). The PDAG is often used as an interpretation of a Markov equivalence class of DAGs which all encode the same conditional independencies. Under the assumption of CBNs, directed arcs in the PDAG represent relationships where the direction of causality is defined, and undirected edges where the direction of causality is unknown.

One of the earliest constraint-based algorithms, which we use in this study, is the Peter and Clark (PC) algorithm (Spirtes and Glymour, 1991) which starts from a fully connected undirected graph. The PC algorithm first checks whether each pair of variables is independent. If they are independent, the algorithm removes the edge between the pair since such an edge would imply dependency. The algorithm then checks conditional independencies between every pair of variables using conditioning sets of increasing size. Edges are removed between nodes if a conditional independence between them given some conditioning set is found. The result is a skeleton of undirected edges – that is, the graph has the correct edges, but none of them are orientated. A search for v-structures is then undertaken, with the v-structures being used to orientate as many edges as possible.

The other constraint-based algorithms used in this study[1] are the Incremental Association (IAMB) (Tsamardinos et al., 2003) and Grow-Shrink (GS) (Margaritis, 2003) algorithms which both use Markov Blankets to first identify the neighbours of each node. The Markov Blanket of a particular node is defined as the conditioning set of nodes which ensure that the particular node is conditionally independent of all other nodes in the graph. Using Markov Blankets reduces the number of conditional independency tests that must be performed making these algorithms faster than the PC algorithm.

A different type of structure learning algorithms are the score-based algorithms, which follow a more traditional machine learning approach that searches the space of possible graphs and returns the graph that maximises a scoring function. The scores used by these algorithms are based around the likelihood of seeing the observed data given the generated graph. The commonly used Bayesian Information Criterion (BIC) score is used in this study to measure this. Scutari and Denis (2014) define the BIC score as:

$$\mathrm{BIC}(G, D) = \sum_{i=1}^{p} \left[ \log \Pr\left(X_i | \Pi_{X_i}\right) - \frac{|\Theta_{X_i}|}{2} \log n \right]$$

where $G$ is the graph, $D$ the observed data, the first term in the summation is the log-likelihood of the observed data given the graph, and the second term penalises graph complexity through $|\Theta_{X_i}|$ which represents the number of free parameters in the conditional probability tables. This formulation proposed by (Schwarz, 1978) and extended by (Haughton, 1988) to cover multinomial DAG models, has been adopted by various structure learning software. It generates negative BIC scores and assumes that a larger BIC score corresponds to a more accurate graphical structure.

Two common score-based approaches include the Hill Climbing (HC) and TABU search algorithms (Russell and Norvig, 2016) which start from an empty graph and use hill climbing (a form of greedy search) to add, remove or re-orientate edges, as determined by a global graph score. The TABU search, which we use in this study, is in fact a HC search with additional search steps that often help to escape local optimal at the expense of higher computational time. The Greedy Equivalent Search (GES) algorithm has also had a major impact in this area of research (Meek 1997; Chickering, 2002). In this study, we use the Fast Greedy Equivalent Search (FGES) which is an optimised and a parallelised version of GES (Ramsey et al., 2017). It starts the search process with an empty graph and progressively adds edges between variables using a forward stepping search in order to increase the BIC score. A backward stepping search is then performed that removes edges until no edge removals increase the BIC score. We also make use of a local-search algorithm called Saiyan which, unlike other algorithms, forces all variables to be part of the same graph under the assumption that all of the data variables are dependent (Constantinou, 2020).

---

[1] It had been intended to include the Fast Causal Inference (FCI) (Spirtes et al., 1999) algorithm but it took too long to run. This was disappointing as it is an interesting algorithm which accounts for the effects of latent confounders (variables not present in the data which are parents of multiple variables in the data).

Other algorithms make use of both constraint-based tests in conjunction with a score-based search to construct BN graphs. These algorithms are called 'Hybrid'. The most well-established hybrid structure learning algorithm, which is also used in this paper, is the Max-Min Hill Climbing (MMHC) algorithm (Tsamardinos, Brown and Aliferis, 2006). In this algorithm, constraints are first used to restrict the possible BNs to search, and then hill climbing is used to obtain the highest scoring network in the reduced search space. We use a more general hybrid algorithm, 2-phase Restricted Maximisation (RSMAX2) (Scutari et al., 2014) where it is possible to specify the algorithm used in the restrict and maximisation phases individually. In this study, the default parameters for rsmax2 are used – the HITON algorithm (Aliferis et al., 2003) is used to restrict the search space, and hill-climbing again used in the maximise score phase.

### 3. Data pre-processing

DHS surveys are conducted approximately every five years in each of the 90 countries covered. This work uses the most recent survey from India undertaken in 2015 and 2016. It is chosen because preventable childhood mortality due to diarrhoea remains high in India (IIPS, 2017) and the Indian survey has the highest number of data instances of any DHS survey; i.e., 259,627 children aged up to five years old.

DHS surveys are conducted at a representative sample of households, 601,059 in the Indian survey. The survey team questions all adults in the household, but there is a more detailed set of questions for women which focuses on their health, and the health of each of their children aged up to 5 years. Well over a thousand pieces of information are collected for each child including information relating to their mother and household. It is this data which is the starting point for this study. Further details of the survey, sampling methodology and overview results can be found in the 2015-16 Indian DHS survey report (IIPS, 2017).

The response data for each DHS survey is available to download for registered users from the DHS Program website (ICF International, no date). Registration involves providing some basic contact details and a legitimate reason for wanting to download the data; i.e., for research purposes. Access is then generally granted and data for a specific country and survey year can be downloaded.

#### 3.1. Data extraction

The raw survey data is provided in two separate text files containing the information relating to the children and households. Each line in the files relates to a single individual or household and has a fixed width format where the value of each variable is encoded, usually as an integer, and located at a particular column position. Separate dictionary files describe: where each variable is to be found in the line; its data type; its allowed values; and an explanation of what the encoded integer values mean (e.g. 1 = male, 2 = female).

Using the standard Pandas data manipulation library (McKinney, 2010), we have written a Python program to read in the dictionary files, merge the data files, decode the data and write out the values of the variables to be modelled. The output is a human-readable tabular format suitable for input to the BN structure learning algorithms. Care is taken that this program can extract and decode different sets of variables reliably and accurately since many different encoding schemes are used for different variables. Even "missing value" is encoded differently from variable to variable.

#### 3.2. Variable Selection

Many BN structure learning algorithms struggle with large numbers of variables because the number of possible graph structures that have to be assessed grows very rapidly with the number of variables.

As Scutari and Denis (2014) explain there are $\frac{1}{2}n(n-1)$ possible edges, or $n(n-1)$ directed arcs in a graph of $n$ variables, which means the number of possible graphs grows super-exponentially with the number of variables. Early experimentation with some of the structure learning algorithms supported the need to reduce the number of variables considerably given that the DHS datasets contain more than 1000 variables. We excluded variables with a high proportion of missing values (over 40%). For example, the variable indicating whether the child lived in a slum was only collected for 1.6% of children surveyed. Based on the literature review, variables judged unlikely to influence the rate of diarrhoea are also excluded. The residual variables from this manual and admittedly subjective process were ranked in terms of most strongly predictive of diarrhoea[2] using WEKA's (Witten et al., 2016) correlation evaluation (Pearson, 1900) and information gain methods (Cover and Thomas, 2012).

Experimentation showed that the majority of the structure learning algorithms of interest would complete in less than six hours if 20-30 variables were chosen. Informed by the WEKA feature selection results, the 28 variables shown in Table 2 are selected for study. These are the 20 most highly ranked by feature selection, plus seven lower ranked ones identified as relevant in the literature. The target variable of whether the child suffered from diarrhoea is included too. Because some of these variables were numeric, correlation tests were also used to discretise the variables. Correlation tests and some subjective judgment was also used to group some states of categorical variables with a high number of states.

**Table 2.** Variables selected for BN modelling.

| Group | Variable Name | Description |
|---|---|---|
| Geographic | • GEO_Region | Indian State household is in |
| Cultural | • CUL_LanguageGroup | Mother's spoken language |
| | • CUL_Religion | Mother's religion |
| Economic | • ECO_WealthQuintile | Wealth quintile of family |
| Household | • HOU_CookingFuel | Type of cooking fuel |
| | • HOU_ModernWallMaterial | Main wall material |
| Knowledge | • KNW_WatchTV | Frequency of watching television |
| Health Services | • SRV_OKAlone | Issue getting medial help: not wanting to go alone |
| | • SRV_Near | Issue getting medial help: distance to facilities |
| Mother | • MTH_MaternalAge | Mother's age in years (15-49) |
| | • MTH_Education | Mother's educational attainment |
| Birth and Delivery | • DEL_SmallBaby | Size of child at birth |
| Child | • CHI_Age | Child's age in years |
| | • CHI_Weight4Height | Child's weight for height specified as standard deviations from mean |
| Breastfeeding | • BF_BottleFeeding | Drank from bottle with nipple yesterday/last night |
| | • BF_EarlyBreastfeeding | When child put to breast – hours or days |
| | • BF_BreastfedMonths | Months of breastfeeding |
| Water, Sanitation and Hygiene (WASH) | • WSH_WaterTreated | Anything done to water to make safe to drink |
| | • WSH_ImprovedWaterSource | Type of drinking water source |
| | • WSH_SafeStoolDisposal | Disposal of youngest child's stools when not using a toilet |
| | • WSH_ImprovedToilet | Type of toilet |
| | • WSH_WashWithAgent | Hand wash cleaning agent observed |
| Family Planning | • FP_BirthsLast5Yrs | Births in last 5 years $(0-6)$ |
| | • FP_ModernMethod | Family planning method |
| Immunisation | • IMM_Measles | Child received measles immunisation |
| | • IMM_VitaminA1 | Child received vitamin A1 |
| | • IMM_Diptheria | Child received diphtheria immunisation |
| Diarrhoea | • DIA_HadDiarrhoea | Whether child had diarrhoea in two weeks preceding survey |

### 3.3. Missing Data Values

Another important consideration is the presence of missing values in the data since most of the structure learning algorithms, including many of those used in this study, do not support datasets with missing values. These missing values follow two patterns.

First, 'structural' patterns in the sense that the missing values arise because of the methodology of the DHS survey. The prime example of this is the fact that information about disease symptoms in children is only collected for living children. Whilst one would naturally want to include disease symptoms in a CBN that models child mortality, the survey design precludes this. As Fig 1a shows, seven variables shown in red were only collected for live children, and two variables related to breastfeeding shown in green were only collected for the most recent birth. Second, 'non-structural' patterns of missing values which are not apparently related to survey design. Whilst some of these missing values might be random (e.g. individual mothers not remembering something about their child) there remains a concern that these missing values might also have a bias; e.g., some information might have been harder to collect from poorer families or particular ethnic groups.

---

[2] The assumption being made here is that those variables most strongly predictive of diarrhoea are the ones most important to include in a CBN, though it could be argued that this might omit variables which, whilst not strongly predictive of the target variable, might have an important influence in a CBN.

A thorough investigation of missing values and their mitigation was not possible. As a result, for the purposes of CBN structure learning, we have examined the following two straightforward approaches:

i.   Imputed missing values using the missForest R library package (Stekhoven and Bühlmann 2011; Stekhoven, 2013). This algorithm is chosen because no parameters need to be specified and it produces an error estimate of the percentage of wrongly classified categorical variables. Fig 1b shows the estimated imputation error indicated by missForest for each variable. Note that some high error rates, around 40-50%, are indicated for several variables so the imputed dataset should be treated with caution. This may echo observations made in Tang and Ishwaran (2017) that missing value imputation is less reliable when variables are weakly correlated, or not missing at random.

ii.  Ignored instances that have missing values for any one of the 28 variables in the data. This reduces the number of instances from 259,627 to 127,787 children.

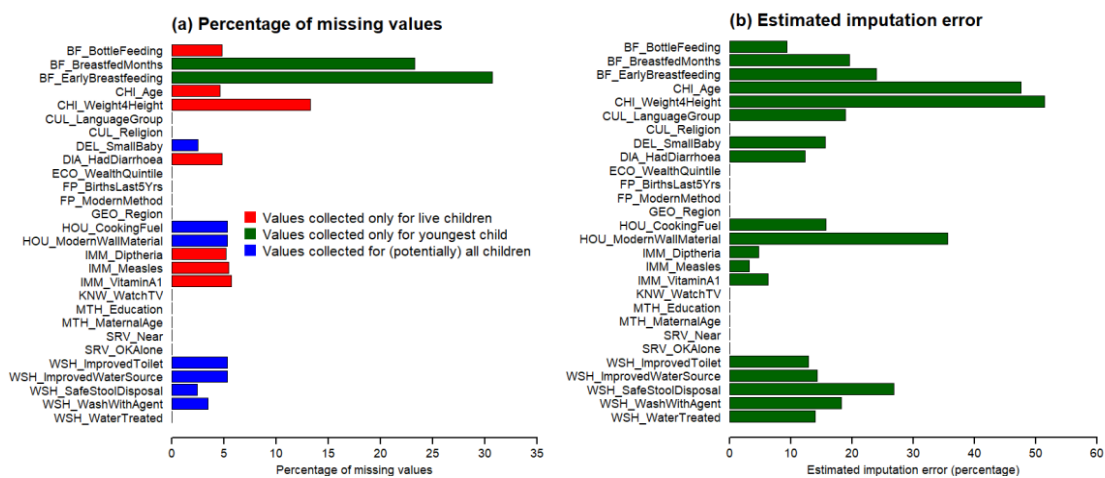

**Fig 1.** Percentage of missing values, along with a possible explanation, for each data variable (left), and estimated imputation error for missing values for each variable (right).

It is important to note that almost all of the structure learning algorithms require complete datasets to work. This means that, in testing these algorithms, missing data points should be imputed or ignored. Other approaches for handling missing values which were not investigated include: a) treating missing values as a distinct category value, and b) the Expectation-Maximisation (EM) approach (Freidman, 1997). The former approach is simple, fast and may be appropriate when the missing values are correlated with other variables (i.e., when missing data points are not missing at random). EM interleaves the graph learning steps with imputation of missing values using the learnt graph at that iteration. However, EM learning needs to be implemented as part of the algorithm, and most algorithms avoid this approach because it is a computationally expensive solution on an even more computationally expensive learning approach.

## 4. Methodology

This section highlights the novel combination of approaches taken to generate and evaluate CBNs for this complex, real-world dataset:

- Use of a wide range of structure learning algorithms provided by three freeware tools.
- Apply background knowledge through the use of synthetic variables and knowledge-based constraints in terms of what can and cannot be discovered by a structure learning algorithm.
- The learned graphs are investigated under different assumptions of missing data points.
- Evaluate the generated graphs with a comprehensive set of metrics.

### 4.1. Tools used for structure learning and parameter estimation

The following freeware BN structure learning tools are used:

i. *TETRAD* is a mature standalone tool supporting a wide range of structure learning algorithms, though predominantly constraint-based, accessed via its own graphical user interface (CMU, 2017). The program was first developed in Pascal in 1986 (Glymour and Scheines, 1986), and has been under continual development since then. New algorithms have been added and it is now implemented in Java. This study uses the latest version available at the time, version 6.5.3, to run experiments with the FGES algorithm.

ii. *Bayesys* is an Alpha release of a tool under active development at Queen Mary University of London. It is implemented in Java and provides an interactive user interface for structure learning and graph evaluation. We used version 1.28 to experiment with the Saiyan algorithm.

iii. *bnlearn* is a software library which is not offered as an application with its own user interface, but provided as a downloadable R package (Scutari and Denis, 2014; Scutari, 2019a). Using the bnlearn package therefore requires some programming in R (R Studio Team, 2019). It provides a wide range of structure learning algorithms which are used in this study: GS, IAMB, MMHC, PC, RSMAX2 and TABU. This study also uses its comprehensive features for parameter estimation and graph evaluation.

Using TETRAD, Bayesys and bnlearn provides a good opportunity to compare and contrast these tools. Table 3 provides an assessment of the tools according to the following criteria identified as being important for this work:

i. *Programmatic access*. The ability to invoke learning and evaluation functions in a programming environment is invaluable, allowing a sequence of tests – perhaps where the data or algorithm parameters are varied - to be run and results recorded automatically. Working in a programming environment also facilitates integration with other packages, e.g. charting or missing value imputation packages.

ii. *Software quality*. In a context where structure learning algorithms take many hours to run, are difficult to set up, and do not save interim results, unpredictable software crashes and infinite loops cause research time to be lost.

iii. A range of different *structure learning algorithms*.

iv. Indication of *structure learning progress*. Some algorithms take several of orders of magnitude longer than others. An indication of how far the algorithm has progressed and an estimated time for completion is therefore very helpful.

v. *Structure learning transparency*: many of the algorithm implementations provide little indication of how the structure they generate arises from the data. A mechanism to convey this information would be useful, ideally in a graphical form such as Tensorboard (Google, 2019) provides for neural network learning.

vi. *Knowledge-based constraints* can be incorporated (discussed in Section 4.2).

vii. *Good graph visualization*: ideally, this should support both automatic and manual layout of generated graphs, and the ability to enter a reference graph manually.

viii. *Graph evaluation*. Supporting the metrics discussed in Section 4.4.

**Table 3.** Evaluation of BN structure learning tools used in this study.

| Feature | TETRAD | Bayesys (Alpha release) | bnlearn + RStudio |
|---|---|---|---|
| Programmatic Access | Yes – Java, R and Python APIs available | None discovered – much time expended running tests manually | Yes |
| Software quality | Poor – unexplained crashes and infinite loops within structure learning and graph layout | Fair – though some bugs encountered handling constraints which are now fixed | Good – no problems encountered |
| Structure learning algorithms | Good - a comprehensive range of algorithms supported | Limited – only a single algorithm supported currently | Good - a comprehensive range of algorithms supported |
| Structure learning progress | No | Fair – percentage completed is reported for the time-consuming first phase | No |
| Structure learning transparency | Poor - incoherent and inconsistent logging messages provide little understanding | Fair – informative visual renditions of the graphs at each phase are produced showing relevant scores on each arc | Fair – debugging can be switched on for most functions producing detailed but very verbose textual logging of the processing |
| Knowledge-based constraints | Good set of graphical features for specifying knowledge constraints including temporal tiers, and required and prohibited edges | Fair – temporal tiers and required edges can be specified via input files | Fair – required and prohibited edges can be specified as input arguments to structure learning functions |
| Graph Visualisation | Good visualization including manual and automated layout and the ability to create and adjust graphs manually | Fair – reasonable renditions of graphs, though no features to create or adjust graphs manually. | Fair – reasonable renditions of graphs, though no features to create or adjust graphs manually. |
| Graph evaluation | Fair – interactive graph description and comparison features provided, though no consistent support for whole graph scores discovered | Fair – a wide range of graph comparison metrics are supported, but not for whole graph scores | Good – a wide range of graph scores and comparison metrics are supported. |

As Table 3 illustrates, none of the individual tools provide all of the desirable features for this study. Initially, TETRAD was used, but its unreliability and opaque behaviour, as well as the necessity of running tests manually, meant experimentation and evaluation was very time-consuming, and management of results difficult. Bayesys on its own would have limited the study to one algorithm. Bnlearn within R Studio is more promising since it allows more automation and provides a wide range of learning algorithms and evaluation approaches. Both Bayesys and TETRAD provide facilities to export the graphs they generate to files. We therefore coded small R functions to read these graphs into R Studio and evaluate them using evaluation functions provided by bnlearn. The only metric computed outside of R Studio and bnlearn is the Balanced Scoring Function provided by Bayesys.

The computational efficiency of the various tools and algorithms is not investigated in any detail beyond simple measurement of the elapsed time to learn a graph. Manual monitoring of multi-core CPU usage does, however, suggest that none of the tools are taking advantage of the multiple cores available. Recent work on parallel learning (Gao and Wei, 2018, Zarebavani et al., 2019) suggests that considerable performance improvements are possible for some algorithms by exploiting parallelism.

### 4.2. Synthetic variables

Synthetic variables are new variables added to the dataset which are deterministically computed from the value of their parent nodes. It is of interest to see whether synthetic nodes can be added to the existing nodes to improve the causal structure of the generated graphs and to evaluate their impact on other graph metrics. Previous work by Constantinou et al. (2016) has noted that synthetic nodes can be helpful for reducing model dimensionality and the effects of combinatorial explosion as well as improving the overall CBN structure of the model in terms of influential relationships. The four synthetic nodes and the parent nodes from which they are computed are summarised in Table 4.

**Table 4.** Synthetic variables incorporated into the structure learning process.

| Synthetic variable | Explanation | Parent variables |
|---|---|---|
| SRV_Accessible | Composite variable indicating whether health services are accessible to the mother as an aggregate of whether they are far away and whether she feels comfortable visiting them alone | SRV_OKAlone<br>SRV_Near |
| BF_GoodBreastfeeding | Composite breastfeeding variable aggregating whether breastfeeding was initiated early, whether bottle feeding was used, and how long breastfeeding was undertaken | BF_EarlyBreastfeeding<br>BF_BottleFeeding<br>BF_BreastfedMonths |
| WSH_GoodWASH | Composite variable indicating the overall level of water, sanitation and hygiene practices found. Includes quality of water source and toilet, whether water treated and a handwashing cleansing agent available, and whether children's stools were disposed of safely. | WSH_ImprovedWaterSource<br>WSH_ImprovedToilet<br>WSH_WaterTreated<br>WSH_WashWithAgent<br>WSH_SafeStoolDisposal |
| IMM_WellProtected | Composite variable reflecting the level of immunization protection comprising whether child was immunized against Diptheria and Measles, and whether given Vitamin A1 supplement | IMM_Diptheria<br>IMM_Measles<br>IMM_VitaminA1 |

### 4.3. Knowledge-based constraints

While this study focuses on machine learning to generate the graphical structure of a CBN model, we are still interested in exploring the impact knowledge-based constraints have on the learning process of the selected algorithms. A fundamental part of our understanding of causality and time is that later events cannot cause earlier events. Hence, we want to prohibit edges from a node to another node which represents an earlier event. Such temporal constraints are used in Constantinou and Fenton (2018) in the context of sports prediction and by Bonchi et al. (2017) when developing a causal model of discrimination. These restrictions are often specified to learning algorithms through the definition of temporal tiers; i.e., variables in tier 1 relate to events earlier than those in tier 2, and so arcs indicating influence from tier 2 (or higher) variables to tier 1 variables are prohibited. This pattern may be repeated over many tiers.

**Table 5.** Temporal constraints incorporated into the structure learning process.

| Tier | Variable/s | Justification |
|---|---|---|
| 1 | GEO_Region | Likely no other variables cause a family to be in a particular region – causality typically the other way around e.g. region → wealth, religion. |
| 2 | MTH_MaternalAge | Mother's age primarily a result of when survey conducted. |
| 3 | CUL_LanguageGroup<br>CUL_Religion | Assume these are pre-determined for a family and only affected by GEO_REGION and no other variables. |
| 4 | ECO_WealthQuintile | Assume this may be an effect of region, ethnic group or religion but not an effect of any other variables. |
| 5 | MTH_Education<br>KNW_WatchTV | The assumption is this is most likely to be affected by socio-economic, cultural and geographic factors. |
| 6 | HOU_CookingFuel<br>HOU_ModernWallMaterial<br>SRV_OKAlone | Assume these are mostly effects of socioeconomic, geographic and cultural factors. |
| 7 | SRV_Near | Assume largely an effect of wealth and geography. |
| 8 | SRV_Accessible | Synthetic node we wish to be an effect of SRV_OKAlone and SRV_Near. |
| 9 | FP_ModernMethod | May be an effect of many of the above variables but temporally precedes variables in lower tiers. |
| 10 | FP_BirthsLast5Yrs | Number of births likely an effect of family planning method. |
| 11 | CHI_Age | Child's age likely affected by family planning and maternal age. |
| 12 | DEL_SmallBaby | Size of baby at birth temporally precedes variables in lower tiers. |

An example of a temporal constraint in this study is that a child's size at birth occurs before an occurrence of diarrhoea and so the latter cannot cause the former. Such clear-cut temporal constraints are uncommon in our dataset, but tiered constraints are also used to prohibit edges for relationships that are judged highly unlikely causally. For example, it is conceivable a family's religion might be a cause of whether they have an improved water source, but the reverse – that having an improved water source is a cause of which religion they adopt – is very unlikely. Table 5 lists the temporal constraints used in our experiments.

A second type of knowledge constraint is also investigated where directed edges are required between some nodes. In particular, the effect of requiring edges from the breastfeeding, immunisation & water, sanitation & hygiene synthetic nodes and the child's weight for height node to the diarrhoea occurrence node is examined. These required edges encourage the structure learning algorithms to create a graph whereby the causes of diarrhoea identified in the literature become antecedents of the diarrhoea node.

### 4.4. Evaluation Metrics

There are two ways to evaluate a BN structure learning or a causal discovery algorithm: a) in terms of how well the learned graph predicts the ground truth graph, and b) in terms of how well the learned distributions fits the empirical distributions. However, there is no agreed approach to evaluate the effectiveness of these algorithms. Because of this, both of these approaches are used in the paper, amongst others, including multiple metrics that are part of each approach. Specifically, we made use of the following metrics:

i. **Comparison with a knowledge-based reference graph:** Evaluations of generated CBN graphs rely on a comparison with a knowledge-based graph or a ground-truth synthetically generated hypothetical graph (Raghu, Poon, and Benos, 2018). The knowledge-based reference graph for this work is shown in Fig 2. Although this graph is created by the first author of this study who has some experience in this domain, it is important to acknowledge that he is not an expert in childhood diarrhoea by training. We compare the generated graphs to the knowledge-based graph using the standard metrics of Recall and Precision (and the resulting F1 score), the SHD score which counts the number of differences between the two graphs (Tsamardinos et al., 2006), and the BSF metric that balances the score proportional to the number of edges in relation to direct independencies to eliminate possible bias in favour of graphs with limited number of edges (Constantinou, 2019).

The graph comparison metrics used throughout are all based on comparing adjacency and direction. As noted previously, the constraint-based algorithms may produce a partially directed acyclic graph. In these cases, undirected edges in the generated PDAG are randomly orientated – if this happens to result in a cycle being created, the opposite orientation is used instead. The PDAGs are thus converted into DAGs allowing their comparison with the reference DAG.

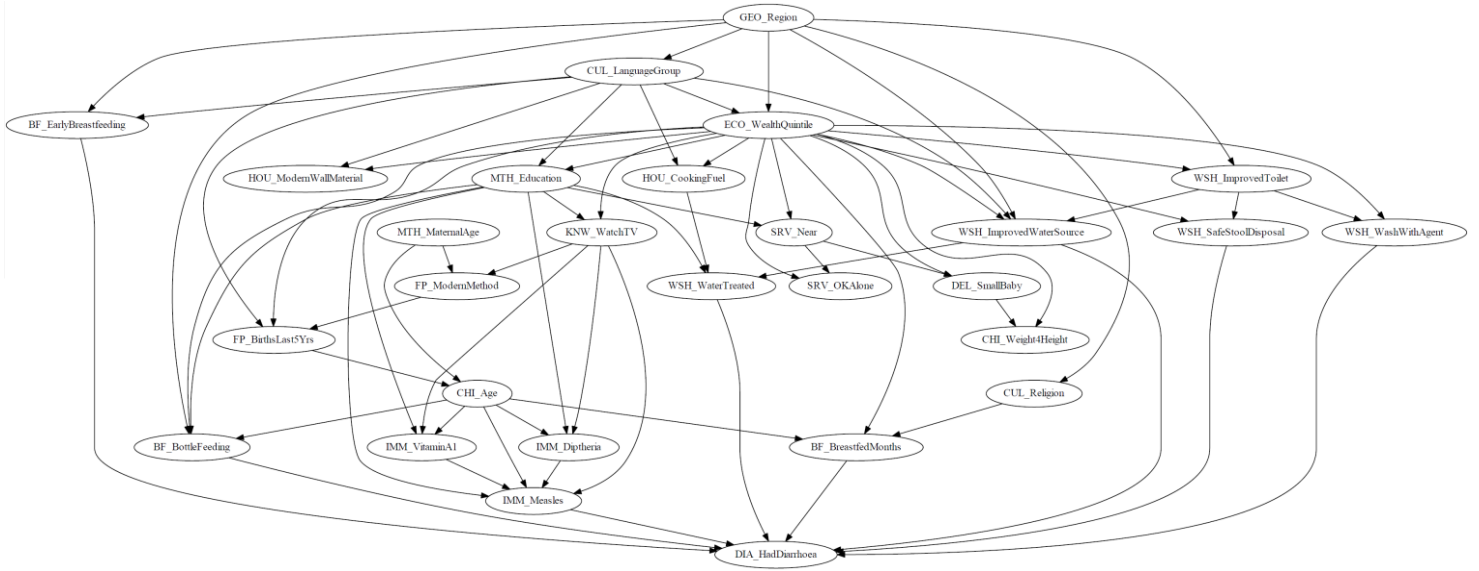

**Figure 2.** The knowledge-based graph used as a reference.

ii. **Graphical properties:** The generated graph's structure is categorised by the number of
    a. Independent (i.e. unconnected) graphical fragments. The variables in this study are assumed to be related to each other, at least indirectly, so a single graphical fragment is desired. The variable selection process makes this assumption very plausible.
    b. Edges as a measure of graph complexity.
    c. "Free" CPT parameters; also known as the independent CPT parameters. Specifically, for variables $X_i, \dots, X_n$ with corresponding parents $U_i, \dots, U_n$, the number of independent parameters for $X_i | U_i$ is $\left(X_i^{\#} - 1\right) U_i^{\#}$ where $\#$ is the number of instantiations. In general, the lower this number is relative to the number of training instances, the lower the risk of overfitting.

iii. **BIC Score:** the BIC score as defined in Section 2, and as measured by bnlearn. It reflects how well the graph fits the data in relation to model complexity and sample size. Note that the BIC score metric is naturally biased in favour of algorithms that maximise it by design.
        Note that bnlearn reports the log-likelihood and the penalty components of the BIC score separately. The outputs suggest that the log-likelihood element of the BIC score is around 150 times larger than the graph complexity penalty; implying that in this study the BIC score closely approximates the log-likelihood score.

iv. **MLE and cross-validation:** bnlearn also offers a cross-validation evaluation feature (Scutari, 2019b). This takes a dataset and splits it up into two portions. One portion is first used to estimate the graph parameters using Maximum Likelihood Estimation. The parameterised graph is then evaluated by computing the log-likelihood of the other (validation) portion. This process is repeated with 10 different estimation and validation portions which is termed 10-fold cross-validation[3].

v. **Number of 'causal paths':** One of our objectives is to generate CBNs that encapsulate the causes of diarrhoea, and this assessment attempts to assess this. It simply counts the number of directed paths in the generated graph that exist out of the eleven variables identified in the literature as causes of diarrhoea, to the diarrhoea occurrence node. These are termed as 'causal paths' and the metric will be an integer between 0 and 11 inclusive.

vi. **Comparison with unconnected and random graphs:** Further to (i) above, the generated graphs are also compared with two naïve baseline graphs; a graph with no edges (denoted as "empty" graph in the results) which represents the case when all variables are independent of each other, and a randomly generated graph using the Melançon algorithm[4] (Melançon and Philippe, 2004). This graph is denoted "random3" in the results. The expectation is that the graphs learnt from the data should have considerably better metrics than these baseline graphs.

vii. **Elapsed time for structure learning:** Structure learning was performed on a relatively old laptop computer which features a dual core Intel i7-4510 2GHz with 8 GB of RAM, running Windows 10. The elapsed time for the structure learning algorithms to complete is recorded as a crude measure of computational efficiency.

---

[3] The bnlearn documentation does not specify exactly what the loss value returned by the cross-validation function represents, but comparisons with the whole graph log-likelihood score indicate it is the average negative log-likelihood value per instance. A higher than expected cross-validation loss (that is, higher than the average per instance log-likelihood using all the training data) would suggest over-fitting is occurring.

[4] The maximum in-degree and out-degree for the algorithm is set to 3 which creates a random graph with a similar number of edges to the generated graphs.

## 5. Results

Structure learning is performed using the algorithms and tools covered in Section 4.1. Each algorithm has its own parameter inputs which influence the output of the generated graph. However, there are no clear guidelines that specify under what circumstances the input parameters need modification, or how to modify them, depending on available data. Because of this, we test the algorithms with their default parameter inputs as set by the structure learning tools. The only exception to this was that the parameter controlling the maximum number of edges connected to any node provided by FGES algorithm is varied. Results are obtained setting this to 3, 4 and unlimited; indicated by the results labelled FGES3, FGES4 and FGES respectively. This parameter is varied because it offers an intuitive and direct way of controlling the number of edges for this algorithm.

### 5.1. Comparing algorithms by their characteristics of the graphs they produce

This results in this section highlight the different characteristics of the graphs produced by the different structure learning algorithms and are based on the dataset with imputed missing values, but without any synthetic variables or knowledge constraints used. Table 6 shows the full set of results from this set of experiments. Particular aspects of these results will be illustrated by figures in the rest of this section.

Fig 3a illustrates the number of edges with respect to the number of independent graphical fragments in the graphs produced by each algorithm. The following patterns can be seen:

i.   GS and RSMAX2 produce graphs with the lowest number of edges (31) and the highest number of independent graphical fragments.
ii.  IAMB, MMHC and PC produce graphs with four to six independent graphical fragments and correspondingly 44 to 54 edges.
iii. Saiyan, FGES/3/4 and TABU are the only algorithms which managed to produce a single graph. However, they achieved this with a highly variable number of edges; from Saiyan that produced 38 edges to TABU that produced 111 edges. We also note how the parameter tweaking (node degree) of FGES exerts a large and expected influence on the number of edges in the graph; i.e., FGES produces 41 edges when its node degree is limited to 3, and 106 edges when no limit is imposed.

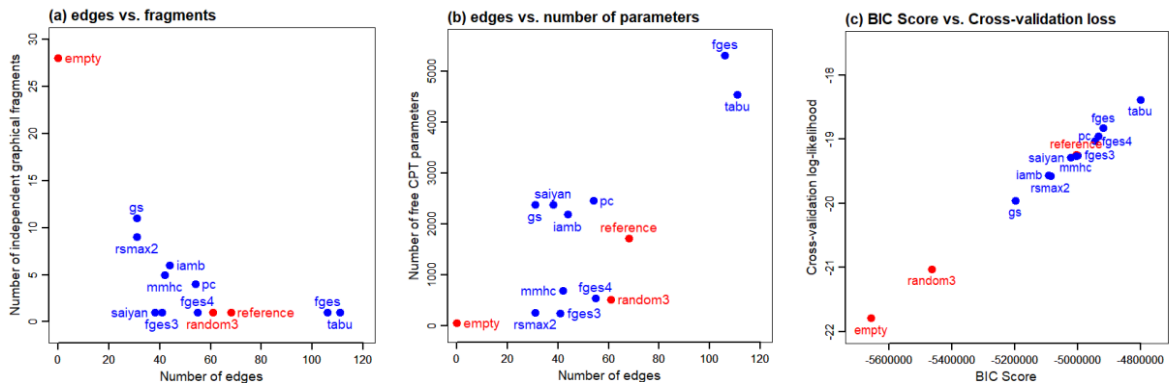

**Figure 3.** Number of edges vs. number of independent graphical fragments produced by each of the algorithms (left), number of edges vs. number of independent graphical fragments produced by each of the algorithms (middle), and BIC scores vs. Cross-validation log-likelihood for each of the algorithms (right).

Fig 3b plots the number of free CPT parameters against the number of edges for each algorithm. As expected, more edges lead to more parents per node, and hence more parameters as evidenced by the trend in this chart. Nonetheless, GS and Saiyan have a relatively large number of parameters compared to the number of edges.

**Table 6.** Properties of the graphs produced by each of the algorithms based on dataset (i); i.e., 28 variables, ~260k samples, and imputation of missing values.

| Graph label used in figures | Algorithm used | Algorithm type | GRAPH DETAILS | | | GRAPH SCORES | | COMPARISON WITH THE REFERENCE GRAPH | | | | | | | | OTHER | |
|---|---|---|---|---|---|---|---|---|---|---|---|---|---|---|---|---|---|
| | | | Independent graphical fragments | Number of edges | Number of free CPT params | BIC score | Cross- validation LL loss | TP | FP | FN | Precision | Recall | F1 | SHD | BSF | Number of causal paths | Time complexity (seconds) |
| empty | Graph with no edges | - | 28 | 0 | 45 | -5657k | -21.79 | 0 | 0 | 68 | 0.000 | 0.000 | 0.000 | 68 | 0 | 0 | 0 |
| fges | FGES [max degree = 100] | Score-based | 1 | 106 | 5315 | -4919k | -18.83 | 29 | 77 | 39 | 0.274 | 0.426 | 0.333 | 124 | 0.294 | 8 | 58.9 |
| fges3 | FGES [max. degree = 3] | Score-based | 1 | 41 | 241 | -4999k | -19.25 | 11 | 30 | 57 | 0.268 | 0.162 | 0.202 | 85 | 0.160 | 7 | 18.3 |
| fges4 | FGES [max. degree = 4] | Score-based | 1 | 55 | 540 | -4944k | -19.03 | 16 | 39 | 52 | 0.291 | 0.235 | 0.260 | 93 | 0.184 | 8 | 19.1 |
| reference | "Best-guess" reference graph | - | 1 | 68 | 1716 | -5005k | -19.24 | 68 | 0 | 0 | 1.000 | 1.000 | 1.000 | 0 | 1 | 11 | 0 |
| gs | GS | Constraint-based | 11 | 31 | 2375 | -5197k | -19.96 | 10 | 21 | 58 | 0.323 | 0.147 | 0.202 | 71 | 0.175 | 3 | 23.5 |
| iamb | IAMB | Constraint-based | 6 | 44 | 2181 | -5091k | -19.56 | 13 | 31 | 55 | 0.295 | 0.191 | 0.232 | 82 | 0.197 | 0 | 12.5 |
| mmhc | MMHC | Hybrid | 5 | 42 | 686 | -5003k | -19.26 | 12 | 30 | 56 | 0.286 | 0.176 | 0.218 | 86 | 0.143 | 0 | 289.8 |
| pc | PC stable | Constraint-based | 4 | 54 | 2456 | -4934k | -18.95 | 15 | 39 | 53 | 0.278 | 0.221 | 0.246 | 89 | 0.211 | 0 | 81335.1 |
| random3 | Randomly generated [max in/out-degree = 3] | - | 1 | 61 | 514 | -5462k | -21.03 | 5 | 56 | 63 | 0.082 | 0.074 | 0.078 | 114 | -0.022 | 1 | 0 |
| rsmax2 | RSMAX2 | Hybrid | 9 | 31 | 252 | -5086k | -19.58 | 9 | 22 | 59 | 0.290 | 0.132 | 0.182 | 73 | 0.135 | 0 | 454.2 |
| saiyan | Saiyan | Score-based | 1 | 38 | 2376 | -5021k | -19.29 | 13 | 25 | 55 | 0.342 | 0.191 | 0.245 | 73 | 0.227 | 0 | 1150 |
| tabu | TABU | Score-based | 1 | 111 | 4539 | -4799k | -18.39 | 34 | 77 | 34 | 0.306 | 0.500 | 0.380 | 113 | 0.379 | 7 | 18.5 |

Fig 3c shows the BIC score and cross-validation log-likelihood loss (the negated expected log-likelihood) for each of the graphs produced. The higher the BIC score and cross-validation log-likelihood the better. The results illustrate the relative superiority of the algorithms with reference to the empty and random graphs. Overall, some algorithms performed worse than the reference graphs, and a similar number performed better. However, on these metrics, the GS algorithm fares substantially worse and TABU substantially better than the rest. As one might expect, the algorithms which explicitly maximise the BIC score by design, such as TABU and FGES, have the highest score. The PC algorithm which does not aim to maximise the BIC score, nonetheless, has performance that is on par to FGES.

The fact that some algorithms (e.g. TABU) produce a higher BIC score than the reference graph is not surprising. This is because the reference graph used in this paper is a knowledge-based graph oriented towards causal representation rather than fitting and hence, some algorithms which focus on maximising a fitting score are likely to achieve a higher fitting-based score. Still, the linear trend in Fig 3c shows that the graph score based on all the instances correlates well with the log-likelihood computed from the cross-validation validation sets. This suggests that these graphs are not over-fitted; if they were, one might expect the validation score to worsen relative to the BIC score for some algorithms.

Fig 4 compares the structure of the generated graphs with the reference graph using precision, recall and F1 metrics. Higher scores indicate stronger similarity to the reference graph. TABU and FGES have the highest recall scores, and have higher recall than precision, possibly because they have the largest number of edges, and so are more likely to match edges in the reference graph. All the other algorithms have higher precision than recall, suggesting they may be too conservative in the number of edges they create for these data. TABU and FGES also have the highest F1 scores, which combine precision and recall, whereas RSMAX2 has the lowest F1 score followed by FGES3 and GS.

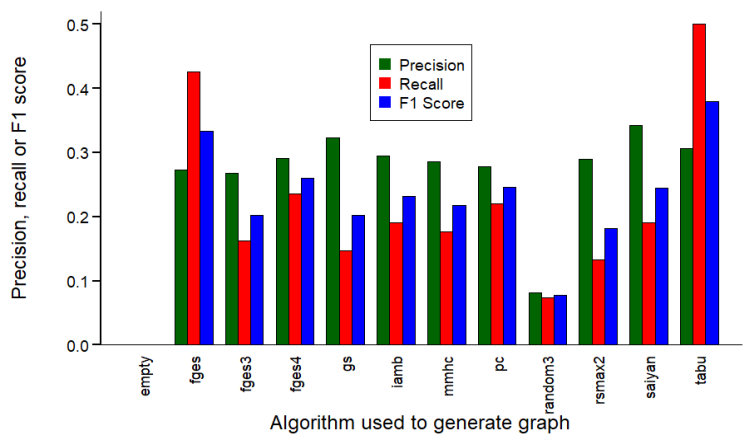

**Figure 4.** Precision, Recall, and F1 score for each of the algorithms, relative to the reference graph.

The SHD scores in Fig 5a provide another comparison metric between the generated graphs and the reference graph. Note that for this metric, *lower* scores indicate stronger similarity between graphs. This score suggests that FGES and TABU are the most dissimilar to the reference graph, which is the completely reverse conclusion compared to the other scores. Indeed, the SHD metric assesses the empty graph as the closest to the reference graph, and even considers the random graph to be on par with TABU and superior to FGES. This outcome supports the suggestion by Constantinou (2019) that the SHD metric tends to be biased towards graphs with low numbers of edges, due to representing pure classification accuracy which can be misleading, and often leads to conclusions that are inconsistent with metrics that offer more balanced scores.

Fig 5b shows the comparison scores using the Balanced Scoring Function (BSF) metric. On this measure, the TABU and FGES perform best, which is in agreement with the Precision, Recall, and F1 metrics, but not with SHD. As would be hoped, all the generated graphs are more similar to the reference graph than the baseline empty and random3 graphs using this metric.

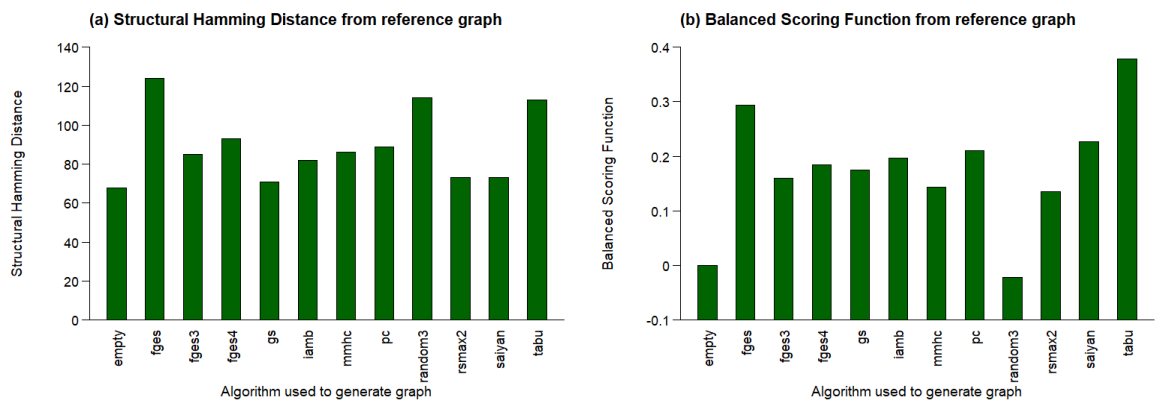

**Figure 5.** Structural Hamming Distance (SHD) scores for each of the algorithms, relating to the reference graph (left), and Balanced Scoring Function (BSF) scores for each of the algorithms, relative to the reference graph (right).

Overall, FGES and TABU produce most edges and are, therefore, more likely to discover true edges. On the other hand, the larger number of generated edges would also include more false edges. Fig 6 illustrates how the graph generated by the TABU algorithm, which contains the most edges, compares to the graph generated by GS which contains the least number of edges. This comparison reveals the complexity of the TABU graph; it is hard to discern the causal structure in this graph even though it tops most of the scores. The figure also shows the large difference between TABU and GS graphs: only 12 edges are exactly the same; 12 edges are reversed; and 87 edges generated by TABU do not exist in the GS graph.

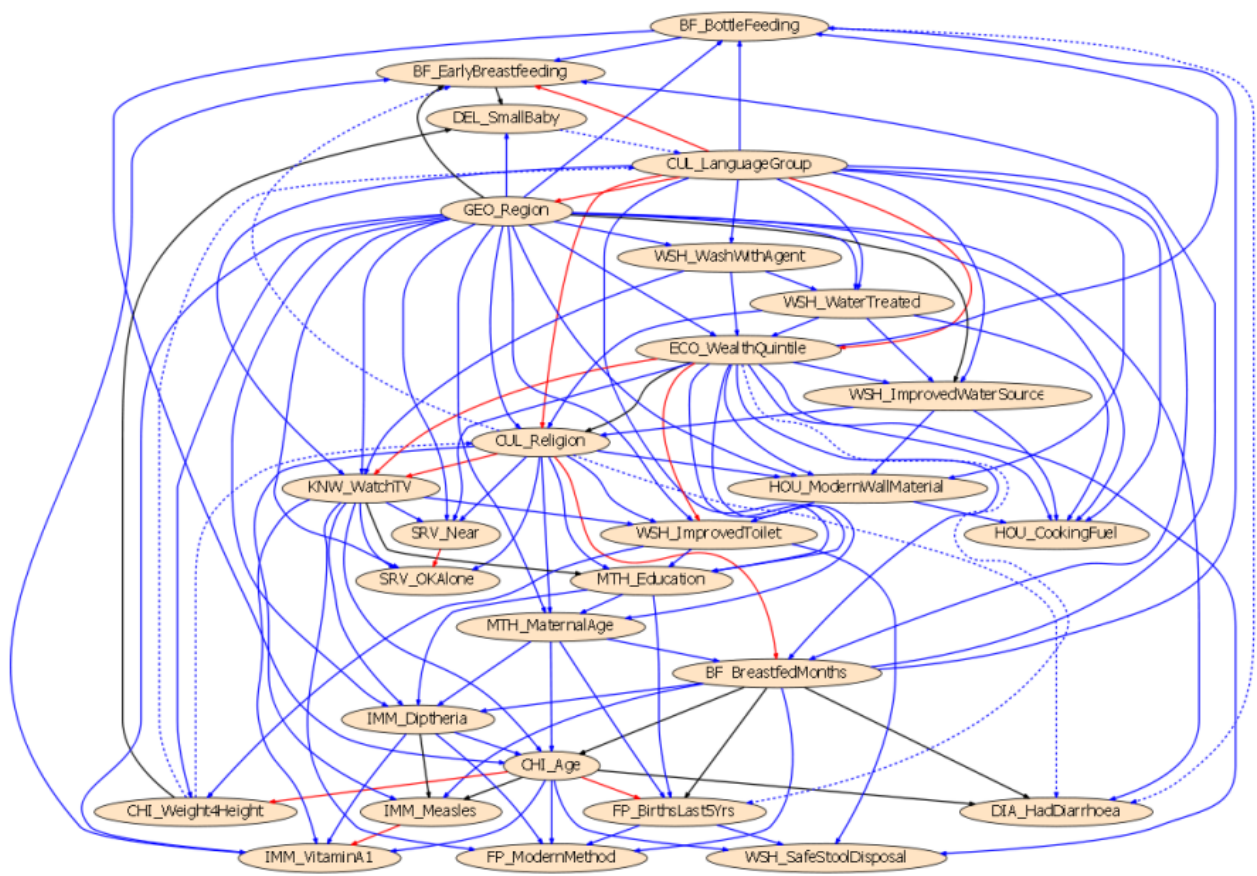

**Figure 6.** TABU graph (contains most edges) compared to the GS graph (contains least edges). Solid blue edges appear exclusively in the TABU graph, dashed blue edges exclusively in the GS graph, black edges appear in both graphs with the same orientation, and red edges appear in both graphs with different orientation.

Table 7 shows the SHD scores between the generated graphs created by each pair of algorithms. The results suggest that MMHC, IAMB, RSMAX2 and PC produce relatively similar graphs. On the other hand, TABU and FGES are different to most other graphs, presumably due to the high number of edges they create. Interestingly, TABU and FGES are relatively close to each other. Finally, while FGES3 and FGES4 are close to each other, they are quite different from FGES and this shows the strong effect of the FGES parameter in controlling the number of edges per node.

**Table 7.** SHD between graphs generated by each pair of algorithms. Deep green cells have the lowest SHD (greater similarity) and light green the highest (least similarity).

|        | fges | fges3 | fges4 | gs  | iamb | mmhc | pc  | rsmax2 | saiyan | tabu |
|--------|------|-------|-------|-----|------|------|-----|--------|--------|------|
| fges   |      | 104   | 102   | 111 | 88   | 84   | 84  | 95     | 100    | 71   |
| fges3  | 104  |       | 41    | 55  | 52   | 43   | 51  | 46     | 53     | 112  |
| fges4  | 102  | 41    |       | 66  | 64   | 51   | 63  | 58     | 62     | 109  |
| gs     | 111  | 55    | 66    |     | 51   | 56   | 65  | 42     | 48     | 108  |
| iamb   | 88   | 52    | 64    | 51  |      | 43   | 61  | 45     | 62     | 100  |
| mmhc   | 84   | 43    | 51    | 56  | 43   |      | 38  | 32     | 55     | 95   |
| pc     | 84   | 51    | 63    | 65  | 61   | 38   |     | 43     | 55     | 97   |
| rsmax2 | 95   | 46    | 58    | 42  | 45   | 32   | 43  |        | 48     | 101  |
| saiyan | 100  | 53    | 62    | 48  | 62   | 55   | 55  | 48     |        | 101  |
| tabu   | 71   | 112   | 109   | 108 | 100  | 95   | 97  | 101    | 101    |      |

Table 8 provides a similar comparison between all algorithms, but this time using the BSF metric. The BSF score is not symmetric since it weights arc comparisons by the number of dependencies and independencies in the reference graph. Thus, in this comparison of pairs of learnt graphs it is affected by which graph we adopt as the "reference" graph and explains why Table 8 is not symmetric. MMHC, IAMB, RSMAX2 and PC have mostly high BSF scores between each other, echoing the SHD comparisons that the graphs they produce are somewhat similar. As with the SHD score, FGES3 and FGES4 are similar to each other but different to FGES, and we also see that FGES and TABU again have a very high similarity. On the other hand, the GS graph is judged as being very different from all the other graphs according to this metric.

**Table 8.** BSF scores between graphs generated by each pair of algorithms. Deep green cells have the highest BSF score (greater similarity) and light green the lowest (least similarity).

|        | fges   | fges3 | fges4 | gs    | iamb  | mmhc  | pc    | rsmax2 | saiyan | tabu  |
|--------|--------|-------|-------|-------|-------|-------|-------|--------|--------|-------|
| fges   |        | 0.160 | 0.205 | 0.207 | 0.280 | 0.305 | 0.362 | 0.218  | 0.193  | 0.690 |
| fges3  | 0.305  |       | 0.627 | 0.181 | 0.374 | 0.450 | 0.429 | 0.306  | 0.300  | 0.321 |
| fges4  | 0.317  | 0.494 |       | 0.150 | 0.268 | 0.405 | 0.377 | 0.248  | 0.266  | 0.325 |
| gs     | -0.045 | 0.156 | 0.131 |       | 0.252 | 0.150 | 0.159 | 0.260  | 0.232  | 0.095 |
| iamb   | 0.434  | 0.327 | 0.208 | 0.224 |       | 0.450 | 0.324 | 0.352  | 0.164  | 0.394 |
| mmhc   | 0.511  | 0.420 | 0.472 | 0.154 | 0.467 |       | 0.590 | 0.502  | 0.268  | 0.590 |
| pc     | 0.508  | 0.328 | 0.357 | 0.152 | 0.291 | 0.494 |       | 0.423  | 0.326  | 0.529 |
| rsmax2 | 0.416  | 0.353 | 0.342 | 0.278 | 0.454 | 0.638 | 0.641 |        | 0.251  | 0.442 |
| saiyan | 0.367  | 0.320 | 0.356 | 0.245 | 0.213 | 0.317 | 0.447 | 0.242  |        | 0.453 |
| tabu   | 0.673  | 0.166 | 0.207 | 0.136 | 0.261 | 0.316 | 0.366 | 0.221  | 0.226  |       |

Returning to Table 6, the algorithms that generate graphs containing the most "causal paths" are TABU, FGES, FGES3 and FGES4. Again, the result for TABU and FGES may have been another side-effect of them producing more edges overall. Less expected, perhaps, is the good performance on this metric of FGES3 and FGES4, where the constraint on node degree keeps the number of edges low and yet they perform well on this measure. The final column in Table 6 presents the elapsed time taken for each algorithm to generate the graph. There is a very large range in this metric which will be explored more in the next section of results.

Lastly, this section also shows that, whilst individual metrics may present a conflicting picture as to which algorithm generates the better graphs, FGES and TABU tend to perform best across all metrics. The variability even amongst those metrics comparing the generated graphs to the reference graph highlights the sensitivity of results to the metrics adopted.

### 5.2. The effect of sample size

This section investigates how sample size influences structure learning in terms of the number of edges in the graph, the BIC score of the resulting graph, and elapsed time. It should be noted that whilst the number of training instances used to train the graph structure is varied, the BIC score is always evaluated on the full dataset of 259,627 instances. This is a more realistic test of the generated structure, and addresses the issue that the BIC score is dependent upon the number of instances used to compute it. The different sample sizes are generated by randomly sub-sampling the full dataset.

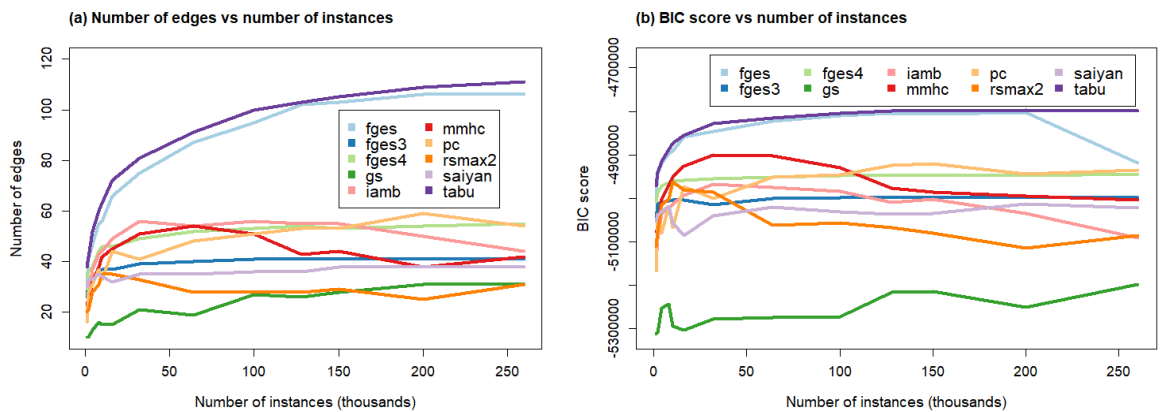

**Figure 7.** How the number of edges discovered by each algorithm is influenced by changes in the sample size (left), and how the BIC score achieved by each algorithm is influenced by changes in the sample size (right).

Fig 7a shows how the number of edges in the generated graph varies with the number of training instances. Though the curves are a little erratic, some patterns clearly emerge. These include:

i.   All of the algorithms show a steep rise in the number of edges as the number of instances is increased from 1000 instances to around 10-20K.
ii.  The FGES and TABU algorithms continue to rise aggressively, though the increase is subject to an exponential decay, with the number of instances. Interestingly, TABU and FGES produce more edges than the other algorithms at all sample sizes.
iii. The FGES3, FGES4, and Saiyan algorithms show virtually no increase in the number of edges beyond 10-20K instances. This reflects the bias these algorithms have to keep the number of edges low; i.e., in FGES3 and FGES4 through the parameter input on the maximum number of parents a node can have, whereas the Saiyan algorithm automatically controls the maximum number of parents a node can have based on the sample size relative to expected number of CPT parameters per variable. However, it should also be noted that Saiyan produces a single graphical fragment, which means that some edges may be 'forced' even in the absence of dependency, and yet the number of edges remain relatively low compared to other algorithms.
iv.  The number of edges produced by IAMB, MMHC and RSMAX2 drop as the number of instances increases beyond 25-50K.
v.   As in the previous section, the GS algorithm behaves rather differently to the other algorithms. It has a more gradual increase, and in nearly all cases produces the lowest number of edges.

The variation of the BIC score with the number of training instances is shown in Fig 7b and follows a similar pattern to the number of edges shown in Fig 7a. The BIC scores rise rapidly for all algorithms as the sample size is increased to around 10-20K. The BIC scores for FGES, PC and TABU continue to rise up to about 150K instances then level off. The modest further increases in edges in TABU and FGES beyond 150K seem not to improve the BIC score further. One peculiarity is the sharp fall in the FGES BIC score between 200K and 259,627 instances; it is unclear why this occurred. The BIC scores for FGES3, FGES4 and Saiyan level off and remain constant, which is consistent with the results in Fig 7a.

On the other hand, the BIC scores for IAMB, MMHC and RSMAX2 fall after around 10-50K instances; implying that the edges eliminated at that point (refer to Fig 7a) have apparently decreased the accuracy of the graph. As with the number of edges, GS continues to differ from the other algorithms, with its BIC score continuing to rise throughout as the number of training instances increases. In the case of PC and the other constraint and hybrid algorithms, the initial increase in number of edges before this point is probably occurring because the conditional independence tests are indicating more nodes as being conditionally dependent as the number of instances grows, leading to a more connected graph. The recent work by Marx and Vreeken (2019) suggests that the conditional independence tests used in this work may indeed be falsely indicating independence with small amounts of data. Investigating the mechanism for this initial rise for the score-based algorithms would be an interesting topic for further work. It would also be useful to explore why IAMB, RSMAX2 and MMHC algorithms see falls in their BIC scores and edge counts beyond around 25K instances.

These results suggest that there is a certain number of training instances required for algorithms to generate consistent graphs. This number varies between algorithms but is usually in the range 10-20K instances for the dataset under assessment. Clearly, sufficient sample size is dependent on the number of variables as well as the number of states per variable in the data.

Finally, Fig 8 illustrates how the elapsed time taken to learn the graph varies between algorithms given the sample size. The elapsed training time for 259,627 instances remains below one minute for all the algorithms except MMHC, RSMAX2, Saiyan and PC. MMHC and RSMAX2 took under 10 minutes for 259,627 instances. Saiyan was the second slowest algorithm beyond 25K training instances. It took around 20 minutes to generate the graph based on 259,627 instances, though elapsed time was close to linear with the number of instances. The elapsed time for the PC algorithm increases exponentially with the number of rows, taking around 23 hours for 259,627 training instances making this algorithm unsuitable for large amounts of data[5].

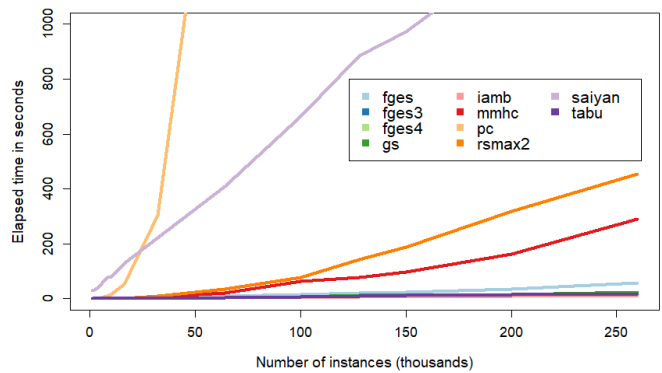

**Figure 8.** How time complexity associated with each algorithm is influenced by changes in the sample size.

### 5.3. The effect of knowledge-based constraints and synthetic nodes

The bar charts in Fig 9 illustrate the effect of knowledge-based constraints and synthetic nodes on the number of edges generated, the BIC score, the BSF score and the 'causal paths' metrics respectively. All these experiments are also based on the 259,627 instances dataset with imputed missing data.

The green and red bars in the charts relate to graphs with 28 nodes, and the blue and orange bars relate to graphs with 32 nodes – that is, with the four synthetic variables added. Whilst comparing graphs with different node sets is unconventional, these results demonstrate the impact of adding deterministic variables which is often adopted when applying BNs to real data.

Fig 9a demonstrates that the application of tiered knowledge constraints alone does not usually alter the number of edges by much, if at all. Examining individual graphs shows that tiered constraints will often only re-orientate related edges. Furthermore, the addition of four synthetic nodes increases the total number of edges by around ten, on average across all algorithms.

---

[5] It was originally intended to study the FCI algorithm. However, it is far slower still – taking four times longer than PC with 11 variables. It was, therefore, impractical to use this algorithm with the 28-variable dataset.

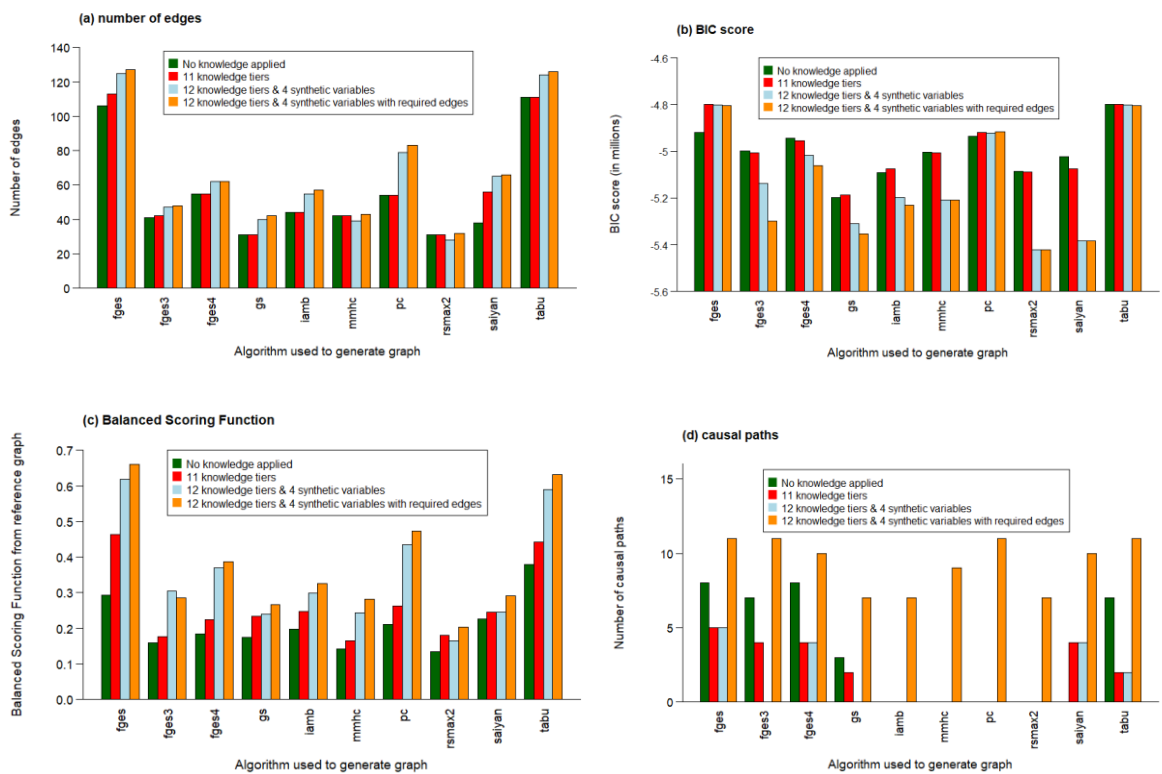

**Figure 9.** A: How the number of edges generated by each algorithm is influenced by knowledge-based constraints and additional synthetic variables, B: how the BIC score generated by each algorithm is influenced by knowledge-based constraints and additional synthetic variables, C: how the BSF score generated by each algorithm is influenced by knowledge-based constraints and additional synthetic variables, and D: How the number of 'causal paths' discovered by each algorithm is influenced by knowledge-based constraints and additional synthetic variables.

Fig 9b repeats the analysis of Fig 9a, but for BIC score rather than the number of edges. Comparing the green and red bars shows that the BIC score is largely unaffected by the addition of constraint tiers. The exceptions to this are Saiyan where the BIC score falls slightly with the addition of tiered constraints, and FGES where it increases considerably. For most algorithms, adding synthetic variables and required edges (light blue and orange bars) worsens the BIC score, especially for the RSMAX2 and Saiyan algorithms. However, the BIC score remains largely unaffected with the FGES, TABU and PC algorithms. A possible explanation is that the FGES, TABU and PC graphs have more edges and hence, these more densely connected graphs can better adapt their structure to accommodate knowledge yet remain more aligned with the observed data. Fig 9c repeats the analysis with reference to the BSF score. This metric shows a consistent picture of the generated graphs moving closer to the reference graph as tiered constraints, then synthetic variables, and finally required edge constraints are added.

Lastly, Fig 9d shows how the constraints affect the number of causal paths. With no knowledge applied, FGES, FGES3, FGES4 and TABU include a significant number of causal paths in their generated graphs. Oddly, adding tiered constraints and synthetic variables reduces the number of causal paths for these algorithms. Saiyan is the sole algorithm that adapts well to these constraints, since the incorporation of tiered constraints and synthetic variables increases the number of causal paths in its graph. However, when tiered constraints, synthetic variables ***and*** required edges are all applied, causal paths appear in all the generated graphs, with TABU, FGES, FGES3 and PC creating graphs with all causal paths present. Note that the required edges are part of the causal paths being assessed and hence, it is not very surprising the causal paths "appeared" in the learned graphs; though it is interesting that this relatively small amount of structure specification can recover the causal paths without great detriment to the BIC score.

Table 9 shows the similarity, as measured by the BSF metric, between the graphs produced by each pair of algorithms when tiered constraints, synthetic variables and required edges are all used. This table may be compared to Table 8 which uses the same colour key but shows these results prior to incorporating any constraints into the structure learning process of the algorithms. Despite the fact that the graphs in the two tables have differing numbers of nodes, the comparison is reasonable given the insensitivity of BSF score to the number of nodes, and provides a reasonable indication as to how some algorithms may benefit, or not, from additional deterministic variables in the data. Pairs of algorithms where BSF similarity has decreased with the application of knowledge constraints are highlighted with red text in Table 9. The constraints have clearly reduced the variation between graphs generated by the different algorithms.

The Saiyan algorithm shows the least convergence with other algorithms when the constraints are applied. A possible explanation may be that its aim of minimising the number of graph edges may clash with the required edges constraint forcing it to adapt differently to other algorithms. Another result of note is the high BSF scores of 0.912 and 0.916 for TABU and FGES respectively, when the full range of knowledge is applied. This is visualised in Fig 10 which show that a large majority of edges are the same in both graphs.

**Table 9.** BSF scores between graphs generated by each pair of algorithms after knowledge-based constraints are incorporated into their structure learning process. Deep green cells have the highest BSF score (greater similarity) and light green the lowest (least similarity).

|        | fges  | fges3 | fges4 | gs    | iamb  | mmhc  | pc    | rsmax2 | saiyan | tabu  |
|--------|-------|-------|-------|-------|-------|-------|-------|--------|--------|-------|
| fges   |       | 0.252 | 0.367 | 0.266 | 0.383 | 0.327 | 0.573 | 0.236  | 0.180  | 0.912 |
| fges3  | 0.541 |       | 0.774 | 0.395 | 0.500 | 0.493 | 0.503 | 0.403  | 0.444  | 0.551 |
| fges4  | 0.634 | 0.619 |       | 0.389 | 0.482 | 0.483 | 0.584 | 0.397  | 0.385  | 0.615 |
| gs     | 0.496 | 0.403 | 0.491 |       | 0.756 | 0.592 | 0.572 | 0.568  | 0.286  | 0.465 |
| iamb   | 0.696 | 0.427 | 0.513 | 0.602 |       | 0.571 | 0.684 | 0.475  | 0.332  | 0.685 |
| mmhc   | 0.714 | 0.527 | 0.643 | 0.592 | 0.716 |       | 0.788 | 0.733  | 0.290  | 0.691 |
| pc     | 0.769 | 0.316 | 0.458 | 0.365 | 0.504 | 0.470 |       | 0.343  | 0.253  | 0.751 |
| rsmax2 | 0.548 | 0.530 | 0.649 | 0.708 | 0.745 | 0.939 | 0.682 |        | 0.295  | 0.564 |
| saiyan | 0.294 | 0.338 | 0.365 | 0.244 | 0.299 | 0.224 | 0.314 | 0.205  |        | 0.304 |
| tabu   | 0.916 | 0.256 | 0.356 | 0.256 | 0.377 | 0.319 | 0.563 | 0.239  | 0.186  |       |

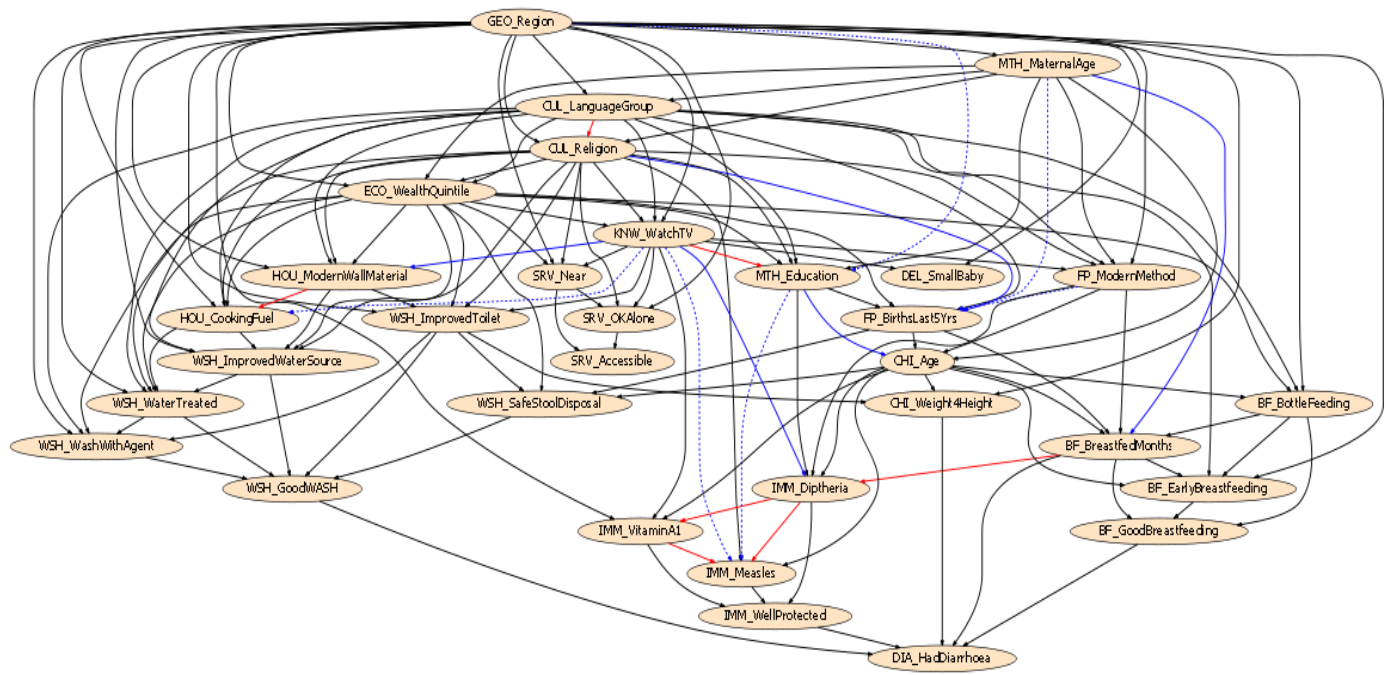

**Figure 10.** Comparison of TABU and FGES graphs after incorporating all of the constraints. Solid blue edges appear exclusively in the TABU graph, dashed blue edges exclusively in the FGES graph, black edges appear in both graphs with the same orientation, and red edges appear in both graphs with different orientation.

We can summarise the diversity between graphs produced across the different algorithms when no knowledge constraints are applied, and when the full range of knowledge constraints is applied, by taking the mean of the BSF score between each pair of algorithms in Table 8 and Table 9 respectively. These are shown in Table 10, together with equivalent comparisons using the SHD and F1 metrics, and scores for when only knowledge tiers are applied. The mean BSF and F1 scores increase markedly as knowledge constraints are applied, indicating that knowledge constraints reduce diversity between graphs as expected – this occurs with and without synthetic variables being added. In contrast, the mean SHD with all knowledge constraints is a little higher than the other two cases, suggesting the opposite conclusion that knowledge constraints **_increase_** diversity. A likely explanation for this contrary result is that synthetic variables add extra edges, tending to increase SHD scores.

**Table 10.** Mean BSF, SHD and F1 scores across all pairs of algorithms, used to compare the diversity of graphs generated by the algorithms, with and without knowledge constraints applied. A higher BSF and F1 scores, and a lower SHD score, indicate stronger similarity between algorithms.

| Metric | Mean score without knowledge-based constraints | Mean score with knowledge tiers only | Mean score with all knowledge-based constraints |
|--------|-----------------------------------------------|--------------------------------------|-------------------------------------------------|
| BSF | 0.331 | 0.374 | 0.448 |
| SHD | 69.1 | 69.2 | 76.4 |
| F1 | 0.281 | 0.410 | 0.497 |

In summary, the results in this section suggest that tiered constraints do not have a large effect on the overall skeleton of a graph under discovery, though they are clearly useful in orientating some edges in the likely causal direction. On the other hand, the addition of synthetic variables and required edges have proven to impose a significant effect on the structure of the graphs and naturally lead to a reduced diversity in the graphs produced by the different algorithms. However, this convergence in graph structure may come at the cost of reduced BIC scores since knowledge-based constraints are at risk of increasing complexity faster than model fitting.

### 5.3. The effect of ignoring missing values

This subsection investigates the effect of treating missing values in two different ways: ignoring rows containing missing values; and, imputing the missing values. Ignoring rows which contained any missing value reduced the dataset size from 259,627 samples to 127,787 samples. We already know, from the results in Section 5.2, that graphs are heavily influenced by sample size. To minimise the risk of having results in this section influenced by sample size, rather than by ignoring missing values, we compare 127,787 instances randomly selected from the 259,627 imputed instances.

Fig 11 compares the graphs generated by each algorithm when missing values are either imputed or ignored. Figure 11(a) shows the BSF score between the imputed and ignored approaches. The BSF scores are relatively high indicating that the approach taken to address missing values may have a lesser effect than the choice of algorithm. The FGES, MMHC, Saiyan and TABU appear to be the least sensitive to the treatment of missing values. Nonetheless, none of the BSF scores is close to 1. In fact, for FGES4, GS and RSMAX2 the BSF scores are below 0.5, which does suggest that for those algorithms missing values do affect the generated graph to quite an extent.

Figure 11(c) shows the precision, recall and F1 scores of the graph where missing values were ignored compared to the graph with imputed missing values – the former being used as the reference graph. This again suggests that while FGES, MMHC, Saiyan and TABU are the least sensitive to the missing value approach, the two approaches examined in this paper produce noticeably different graphs; implying that the selected approach has a strong impact on the learned graphs. Figure 11(b) shows the SHD between the two missing value treatments. According to the SHD measure, the GS and Saiyan algorithms are least sensitive to the missing value treatment which, once more, contradicts the results obtained by BSF and F1 measures.

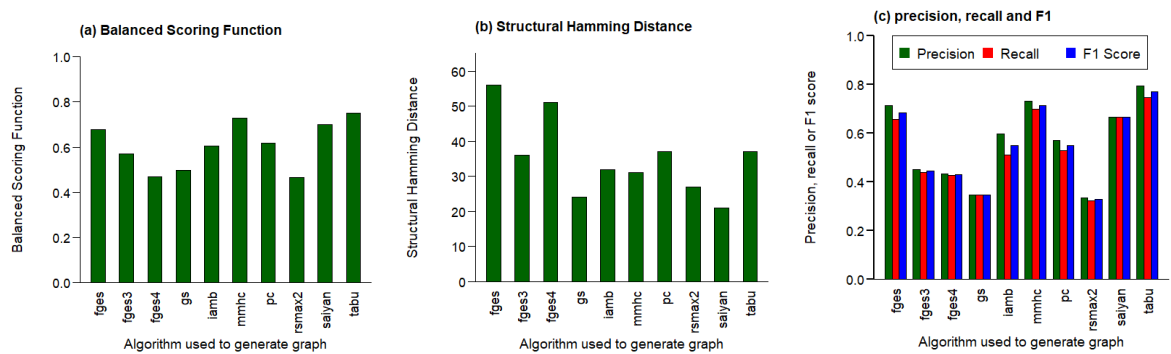

**Figure 11.** Comparisons between graphs generated by the same algorithm but with missing values imputed or ignored as measured by (a) BSF, (b) SHD and (c) Precision, Recall and F1 score.

## 6. Concluding Remarks

The key aim of this study was to examine whether reasonably accurate CBN models can be automatically constructed from DHS data, with and without human knowledge being incorporated. A reasonably accurate CBN graph must provide a basis for predicting the outcomes of preventative interventions by capturing the possible causes of childhood diarrhoea. Papers of similar scope that evaluate BN structure learning algorithms tend to focus exclusively on simulated data and hence, this work provides a much-needed study of a complex real-world dataset. Moreover, this study offers methodological contributions on a key health dataset, and it is reasonable to assume that these contributions have general applicability to other health datasets of similar nature. These include a) the feature selection process on a health dataset that includes repetitive, redundant and contradictory information, b) the use of different structure learning algorithms, that come from different structure learning freeware tools, on determining causal relationships from complex health data, c) the different treatments of missing data points and their effect on structure learning performance, and d) the different approaches to incorporating knowledge-based constraints and their effect on structure learning performance.

Overall, the results from this work suggests that, at the moment, structure learning alone (i.e., without knowledge-based constraints) may be impractical in determining a CBN that provides a reasonably accurate causal representation on the outcomes of preventative interventions on childhood diarrhoea for, at least, two reasons:

i. The DHS survey design and data collection difficulties mean that key information is not collected for some or all of the children. This creates issues with missing values, but perhaps more fundamentally, means that there may be many latent confounders making causal discovery problematic.

ii. The CBNs that are learnt from the data are sensitive to the algorithm used, the number of training instances and the treatment of missing variables. This suggests that the outcome of studies like this may still be rather dependent on the approach adopted and that, perhaps, current methodologies are not yet mature enough to be used with real world survey data such as DHS.

Moreover, as there have been fairly limited attempts in the literature to learn CBNs with health survey datasets of this size and complexity, it is of interest that well-known algorithms struggled with the data in terms of accuracy, consistency and computational performance.

This in turn means that this work is not yet able to offer any new insights into the causal factors behind childhood diarrhoea. The implications of this are that the causes of childhood diarrhoea are more complex than initially thought, both in terms of quantity as well as the relationships between the potential causes of diarrhoea. Whilst there are large variations in behaviour across the algorithms explored, it is possible to discern some groups of algorithms which behave similarly on the DHS data:

i.  FGES and TABU are score-based algorithms and can be characterised by producing a single graph (i.e., includes all data variables) when sufficient data are available, have large numbers of edges, and are the most similar to the knowledge-based reference graph. They are fast, respond well to the incorporation of knowledge, and are the least sensitive to the treatment chosen for missing values. The good performance for TABU echoes that reported by Constantinou (2020).

ii.  IAMB, MMHC, PC and RSMAX2 represent another, but less homogeneous group. They typically produce graphs with several independent graphical fragments (i.e., not all data variables are part of a single graph) and a moderate number of edges which drops as the number of training instances rises above 100K (for the DHS dataset). The BIC score drops as knowledge-based constraints are applied. Elapsed time varies hugely across this group; from IAMB being the fastest to PC being the slowest in this study, supporting the need for algorithms that minimise the number of conditional independence tests as discussed by Gandhi, Bromberg, and Margaritis (2008).

iii.  The GS algorithm seems to have its own character. It produces a large number of independent graphical fragments and correspondingly low numbers of edge, given a dataset that we already know the variables are dependent. It is very dissimilar to the other algorithms according to the BSF metric. It also behaves differently as the number of data instances is changed and it is highly sensitive to the missing value approach taken.

iv.  Saiyan is another algorithm that stands apart which could be attributed to the local-search heuristic and novel assumptions on which it is based. It is designed to produce a single graph under the assumption that all of the data variables are dependent. It generally produces low number of edges and it is less sensitive to the missing value approach. Applying knowledge constraints brings the graph closer to the reference graph.

Despite the relative differences in the sensitivity of the algorithms to the missing value approach, the overall results suggest that the treatment of missing values has a strong impact on the learned graph. Some algorithms struggle with the amounts of data used in this study – particularly PC and FCI – and so seem unsuited to "big data" applications. However, other algorithms, notably FGES and TABU, scale well and look like good candidates for dealing with much larger amounts of data; for example, a larger set of DHS variables. Although we cannot be definitive about the best algorithms to use with this data, the better performance of TABU and FGES over most metrics lead us to recommend them for further work with DHS datasets.

The application of knowledge constraints is found to be beneficial both in terms of reducing the diversity of the graphs produced by the different algorithms, as well as producing a graph which is more reflective of the likely causal relationships in the data. Prediction of childhood diarrhoea rates was not within the scope of this study, but convergence of the graphs produced through the application of knowledge suggests this might be a fruitful topic for further work.

The methodology presented here can be applied to any of the more than 400 DHS datasets covering 90 countries and can be easily adapted to look at other causes of child and maternal death such as pneumonia or malaria. It therefore provides a platform for investigating the variability of causal factors of child and maternal health across time and geographies.

This study also provides insights into what might constitute an ideal tool for exploring Bayesian Networks further; e.g., visualisation of the learning process and a programmatic interface to the learning algorithms would be two recommendations.

To conclude, there seems to be a good opportunity to build on the methodology used but with more focus on the recommended algorithms identified in this work. This, together with improved a) variable selection, b) treatment of missing values and c) use of synthetic variables and knowledge-based constraints, may move us closer to being better able to model the causes of diarrhoea based on the DHS data and reach the goal of predicting the impact of relevant interventions available to decision makers.

## Acknowledgements

This research was supported by the ERSRC UKRI Fellowship project EP/S001646/1 on *Bayesian Artificial Intelligence for Decision Making under Uncertainty*, by The Alan Turing Institute in the UK under the EPSRC grant EP/N510129/1, and by OneWorld UK.